# HEval: Yet Another Human Evaluation Metric

Nisheeth Joshi[1], Iti Mathur[2], Hemant Darbari[3] and Ajai Kumar[4]

[1,2]Department of Computer Science, Banasthali University, India

[3,4]Center for Development of Advanced Computing, Pune, Maharashtra, India

### ABSTRACT

*Machine translation evaluation is a very important activity in machine translation development. Automatic evaluation metrics proposed in literature are inadequate as they require one or more human reference translations to compare them with output produced by machine translation. This does not always give accurate results as a text can have several different translations. Human evaluation metrics, on the other hand, lacks inter-annotator agreement and repeatability. In this paper we have proposed a new human evaluation metric which addresses these issues. Moreover this metric also provides solid grounds for making sound assumptions on the quality of the text produced by a machine translation.*

## 1. INTRODUCTION

Machine translation is the process of converting the text from one natural language into another natural language. In order to analyze the quality of machine translation system we need to evaluate its performance in terms of the translation being produced by the system. This is done by evaluating translations through human or automatic evaluation metrics. In the context of English-Hindi MT systems, automatic evaluation metrics being proposed are highly inappropriate to identify the quality of MT systems. This fact has been shown by several researchers. Ananthakrishnan et al. [1] showed the problems that BLEU MT Evaluation metric [2] faces while evaluating free word order languages. Chatterjee et al. [3] proposed a method through which they showed how a BLEU metric can be modified to evaluate MT systems which have Hindi as a target language.

Since the advent of metrics comparing the translation at deeper linguistic levels, BLEU lost its converted place in evaluation paradigm. Joshi et al.[4] compared results of MT engines between human evaluation, BLEU and Metoer[5] evaluation metrics. They concluded that no evaluation metric, working at different linguistic levels, is capable to capture correct MT quality. This leaves us on the mercy of human evaluation which is also not free from problems. Inter-annotator agreement is one such issue i.e. if a translation is given to multiple judges for evaluation then no two judges would give the same score to that translation leading to confusion where the MT development team would not be able to identify the quality of the system. This problem occurs because human evaluation is inherently subjective. Another problem that the development team faces is the lack of justification being provided as to why certain translation has been termed as good or bad. Keeping in view of these two problems we have proposed a new human evaluation metric which at one end reduces inter-annotator agreement and on the other had provides answers as to why a translation was termed as good or bad.

The rest of the paper is organized as follows: Section 2 gives an overview of some of the popular human evaluation methods being used for analyzing MT quality. Section 3 describes the experimental settings that were done in order to study the performance of this metric. Section 4 shows the working of our evaluation method and section 5 shows the evaluation and results of the metric. Section 6 concludes the paper.

## 2. LITERATURE SURVEY

ALPAC[6] did the first evaluation of machine translation. They incorporated human evaluation were they asked the evaluators to adjudge intelligibility and fidelity of the translations. They randomly took translations of 144 sentences from a Russian book, without knowing which was done by a human and which was done by the MT system.

A major breakthrough in MT evaluation was done by Slype[7] where he evaluated SYSTRAN system. Instead of looking for correctness of the translation, he adjudged SYSTRAN for acceptability. Here the evaluators where asked if translation A is better than translation B. The prime objective of this evaluation was to distinguish between correct sentences from incorrect ones. This evaluation, not only gave a measure to check the output of the system, but also found the cost of post editing the incorrect translations. This evaluation changed the view of the people towards MT evaluation. Now, people started looking at the cost of post-editing translations instead of looking for correct translations. In 1980, a more detailed evaluation was carried out for English-French TAUM-Aviation MT system[8]. Here raw MT outputs and post-edited MT outputs were compared with human translations and the results were analyzed in terms of quality and cost of producing such translations.

Church and Hovy[9] looked at measuring informativeness of the translation. They directly compared the MT systems onto the results of comprehension tests where human evaluators where ask to read MT outputs and then answers certain multiple choice questions. They argument was that if if the translations can capture the information correctly than the user must be able to answer certain questions based on this information.

In 1992, DARPA compared MT system outputs using a comprehension test for intelligibility and a quality test for fidelity[10]. They took passages from various texts as source for translation. They analyzed that this was a very complex and highly expensive method of evaluation, thus in subsequent years they simplified comprehension test. Moreover, the quality test was replaced with adequacy and fluency tests which were assessed on a scale of 1-5.

A more general evaluation methodology was implemented by the Expert Advisory Group on Language Engineering Standards (EAGLES), which started in 1993[11]. Their framework primarily focused on usability evaluations and considered the use of formal descriptions of users, systems, methods and measures along with suitable test materials and where possible some automated evaluation and reporting procedures. The framework distinguished between three types of evaluations:

1. Adequacy evaluation: where the evaluation is done from the perspective of the end user.
2. Progress evaluation: where the evaluation is done to measure the performance of the current version against its previous versions.
3. Diagnostic evaluation: where the focus is on identifying the weaknesses or errors of the system.

Eck and Hori[12] suggested a measure to compare the meaning of the translation with the source. Their arguments what that if the meaning of the source sentence is preserved in the target sentence than it can be considered as a good translation. They rated the translations on a scale of 0-4. Table 1 shows the description of these scales.

Vanni and Miller[13] used clarity as a measure of ascertaining MT quality. They asked the human evaluators to score MT outputs on the basis of the clarity of the translation on a scale of 0-3. Table 2 shows the description of these scales.

| Score | Description |
|-------|-------------|
| 4 | Exactly the same meaning |
| 3 | Almost the same meaning |
| 2 | Partially the same meaning and no new information |
| 1 | Partially the same meaning but misleading information is introduced |
| 0 | Totally different meaning |

Table 1: Meaning Maintenance Scores

| Score | Description |
|-------|-------------|
| 3 | Meaning of the sentence is perfectly clear on first reading |
| 2 | Meaning of sentence is clear only after some reflection |
| 1 | Some, although not all, meaning is able to be gleaned from the sentence with some effort |
| 0 | Meaning of the sentence is not apparent, even after some reflection |

Table 2: Interpretation of Clarity Scores

**What was said:**
> The building has been designed to harvest all rainwater and re-charge ground water using traditional storage system.

**What was translated:**
> भवन और र आरोप आधार पानी पारंपरिक संचयन प्रणाली प्रयोग करके सभी बारिश का पानी फसल एकत्र करने के लिए रूपरेखा प्रस्तुत किया गया है

Table 3: Source Sentence and Target Translation Provided to Bilingual Human Evaluator

| Score | Description |
|-------|-------------|
| 3 | Completely adequate |
| 2 | Somewhat adequate |
| 1 | Tending towards adequate |
| 0 | Neutral |
| -1 | Tending towards inadequate |
| -2 | Somewhat inadequate |
| -3 | Tending towards inadequate |

Table 4: Semantic Adequacy Scores

Gates[14] and Nubel[15] employed semantic adequacy for analyzing MT outputs. They used bilingual human evaluators and asked them to read the source sentence and the corresponding target translation. Based on their understanding of the source and target sentences, they asked them to rate the translations on a multi-point scale. Table 3 illustrates an example of this type of evaluation. Here the evaluator was not only required to rate semantic adequacy, but also was required to rate understandability and fluency of the translation. Table 4 shows the description of this multi-point scale.

Wilks[16] employed monolingual judges and provided them only with translations of source sentences, produced by MT systems. He showed that fluent MT output is also mostly correct and coveys meaning of the source text. Therefore, fluency assessments by a monolingual judge often correlate with semantic adequacy assessments. Thus, he concluded that fluency judgments have some power to predict ratings of semantic adequacy.

Callison-Burch et al.[17] used monolingual judges to rate the informativeness of the translations produced by MT Engines. Here, the human evaluator was only provided with target translation produced by MT Engine and was asked to make out what is the message incorporated in the sentence. Once this was done, the judge was shown human reference translation for the same sentence and was asked to rate how informative or uninformative the translation was.

Doherty et al.[18] employed eye tracking as the measure to evaluate MT systems. They also used monolingual human judges to measure comprehensibility of a translation. They found out that by registering gaze time and fixation count to measure, we analyze if the translation is good or bad as both gaze time and fixation count is more with bad translations then for good translations.

| |
|---|
| ' The Garden [VP] ||| ' द गार्डन [X] ||| 1 0.000963698 1 2.52536e-05 2.718 ||| 0.333333 0.333333 |
| ' Victoria [VP] ||| विक्टोरिया मेमोरियल ' [X] ||| 0.5 2.45014e-06 0.142857 3.86883e-10 2.718 ||| ||| 2 7 |
| Archaeological evidences [NPB] ||| पुरातात्विक प्रमाण [X] ||| 1 4.83131e-07 1 0.00892857 2.718 ||| ||| 1 1 |
| A mountaineer [NPB] ||| एक पर्वतारोही को [X] ||| 1 1.04081e-05 0.5 2.2196e-05 2.718 ||| ||| 1 2 |
| A pond [NPB] ||| एक तालाब [X] ||| 0.5 0.000279033 1 0.00326876 2.718 ||| ||| 2 1 |
| A sheet of golden sunlight [NP-A] ||| सुनहरी धूप की चादर बिछी [X] ||| 1 3.10595e-09 1 6.19203e-10 2.718 ||| ||| 1 1 |
| About 28 [QP] ||| लगभग 28 [X] ||| 0.333333 0.00296589 1 0.0304054 2.718 ||| ||| 3 1 |
| About [IN] ||| के बारे में [X] ||| 0.0833333 0.010989 0.0588235 1.86159e-05 2.718 ||| ||| 12 17 |
| According [PP][X] the slope [PP][X] [VP] ||| के [PP][X] बर्फ की [PP][X] [X] ||| 0.5 1.66449e-07 1 2.18268e-05 2.718 ||| 4-1 1-4 ||| 0.5 0.25 |
| According [VBG] ||| अनुसार [X] ||| 0.130435 0.0756303 0.06 0.0633803 2.718 ||| ||| 23 50 |
| Tourists [VP][X] [NP-A] ||| [VP][X] पर्यटकों को [X] ||| 1 0.0714286 0.0362318 1.17784e-07 2.718 ||| 1-1 ||| 0.125 3.45 |
| Tourists coming to Bandhavgarh [PP][X] [NP-A] ||| शहडोल , उमरिया से बांधवगढ़ [PP][X] ' [X] ||| 1 1.38521e-07 0.0399999 5.1735e-14 2.718 ||| 4-5 ||| 0.1 2.5 |
| moonlit [JJ] ||| चाँदनी [X] ||| 0.333333 0.111111 1 0.4 2.718 ||| ||| 3 1 |
| moonlit nights [NPB] ||| चाँदनी रात [X] ||| 1 0.000793655 1 0.0266667 2.718 ||| ||| 1 1 |
| narrow [JJ] ||| सँकरा [X] ||| 1 0.142857 0.125 0.025 2.718 ||| ||| 1 8 |
| seen in [NPB][X] [VP-A] ||| को पायरटन [NPB][X] में देखा जा सकता [X] ||| 0.476191 0.000223483 0.0262767 2.59799e-16 2.718 ||| 2-2 ||| 0.190909 3.45968 |
| several view points [NPB] ||| अनेक व्यू प्वाइंट [X] ||| 1 5.32722e-05 1 3.73744e-05 2.718 ||| ||| 1 1 |
| should keep in mind [VP] ||| को स्मरण रखनी चाहिए [X] ||| 1 0.000831152 1 3.29656e-07 2.718 ||| ||| 0.333333 0.333333 |
| should not do mountaineering . [VP] ||| पर्वतारोहण किया ही न जाए । [X] ||| 1 3.62662e-10 1 4.26622e-12 2.718 ||| ||| 1 1 |
| zoo [NN] ||| चिड़ियाघर [X] ||| 0.125 0.0291971 0.133333 0.133333 2.718 ||| ||| 16 15 |

Table 5: Snapshot of Hireo Grammar for English-Hindi Tree to String Model

## 3. EXPERIMENTAL SETUP

In order to analyze the authenticity of our metric, we used two most popular MT systems that are freely available on the web. We used Microsoft Translator being developed by Microsoft Corporation and Google Translator being developed by Google Inc. Moreover we also download Moses toolkit[19] and trained two models using this toolkit. One was the baseline system that used phrase based translation model and another was Syntax based translation model which used Tree to String model for training MT system. For syntax based model, we used Carniak parser to generate English parse trees which were then aligned with Hindi sentences. A snapshot of this model is shown in table 5. This type of model, where at one end we have a parse tree and on the other end we have a string, is termed as a hireo grammar. We also used an example based MT system[20] [21] that we had developed to understand the modalities of EBMT and later used it as a Translation Memory.

For training these three systems (two systems through Moses Toolkit and the third one that we had developed), we used 15,000 sentences from tourism domain. The statistics of this corpus is shown in table 6. For tuning these systems, we used 3,300 sentences that were developed for ACL's 2005 workshop on Building and Using Parallel Text: Data Driven Machine Translation and Beyond [22]. The statistics of the corpus is given in table 7. For testing the systems we used 1300 sentences that we had collected separate from 15000 sentences that were used for training the system. We carefully selected these 1300 sentences which capture almost all the aspects of English language. Statistics of this corpus is shown in table 8 and table 9 shows 13 constructs used in the test corpus which capture all the major English aspects.

| Corpus | English-Hindi Parallel Corpus | |
|---|---|---|
| Sentences | 15,000 | |
| Words | 3,34,636 | 3,14,006 |
| Unique Words | 66,788 | 73,066 |

**Table 6:** Statistics of training corpus used

| Corpus | English-Hindi Parallel Corpus | |
|---|---|---|
| Sentences | 3,300 | |
| Words | 55,014 | 67,101 |
| Unique Words | 8,956 | 10,502 |

**Table 7:** Statistics of tuning corpus used

| Corpus | English Corpus |
|---|---|
| Sentences | 1,300 |
| Words | 26,724 |
| Unique Words | 3,515 |

**Table 8:** Statistics of test corpus used

| S.No. | Construct | S.No. | Construct |
|-------|-----------|-------|-----------|
| 1. | Simple Construct | 8. | Coordinate Construct |
| 2. | Infinitive Construct | 9. | Copula |
| 3. | Gerund Construct | 10. | Wh Structure |
| 4. | Participle Construct | 11. | That Clause |
| 5. | Appositional Construct | 12. | Relative Clause |
| 6. | Initial Adverb | 13. | Discourse Construct |
| 7. | Initial PP | | |

**Table 9:** Constructs used in Test Corpus

| Engine No. | Engine Description |
|------------|--------------------|
| E1 | Microsoft Translator |
| E2 | Google Translator |
| E3 | Moses Syntax Based MT |
| E4 | Moses Phrase Based MT |
| E5 | Example Based MT |

**Table 10:** Machine Translators Used

For the rest of the paper we shall be using the five MT engines used by their engine nos. This is shown in Table 10. We have used two human judges to analyze the outputs of MT Engines. This would also help us in checking the inter-annotator agreement.

## 4. HEVAL: HUMAN EVALUATION METRIC

Human evaluation is a key component in any MT evaluation process. This kind of evaluation acts as a reference key to automatic evaluation process. The automatic metrics are evaluated by how well they are able to correlate with human assessments. In a human evaluation, human evaluators look at the output and judge them by hand to check whether it is correct or not. Bilingual human evaluators who can understand both input and output are the best qualified judges for the task. Figure 1 shows the process of human evaluation.

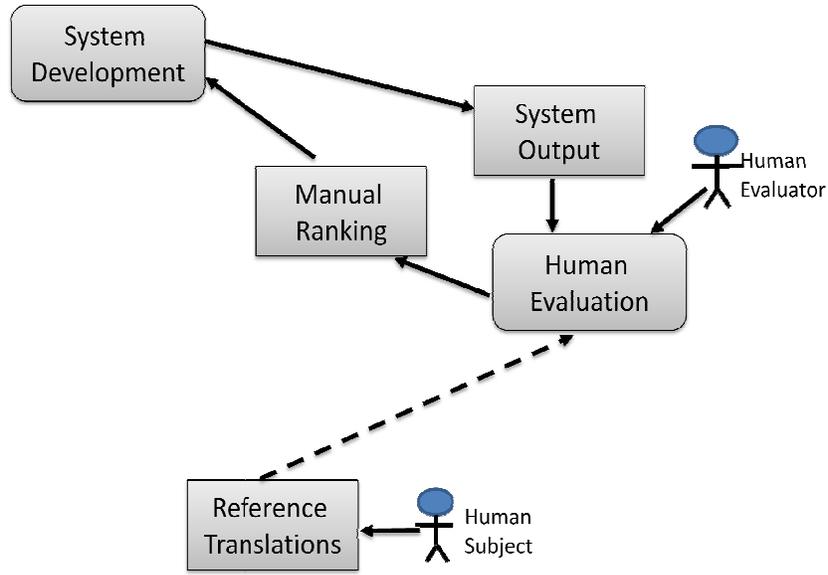

Figure 1: Human Evaluation Process

Here, the output of the system is provided to the human evaluator, who evaluates the output on the basis of a subjective questionnaire/metric, based on which an evaluator can judge the output. This is done for each of the sentences which are going to be evaluated. But, judging MT output merely on the basis of correctness is a very harsh method of evaluation. So, MT Researchers have developed metrics onto which an output can be checked. Most of the human evaluation campaigns judge MT outputs based on either comprehensibility or adequacy or fluency or meaning maintenance or clarity. In recent years, almost all the evaluation campaigns have been using adequacy and fluency as a measure to register human judgments. These two factors can be incorporated in a single metric or can be provided as a separate metrics. Unfortunately their lack of informativeness has lead to various problems. Thus a need is being felt for a metric which can be subjective, effective, consistent and informative at the same time. Here, we shall describe a novel human evaluation metric that we have developed and evaluated the results of five MT Engines. We shall also compare this metric with Human Adequacy and Fluency using Scale 5 metrics and would analyze its results. This human evaluation metric captures the linguistic features of a language and provides qualitative assessments of the MT engine outputs. The linguistic features that we have considered in the metric are:

1. Translation of Gender and Number of the Noun(s).
2. Translation of tense in the sentence.
3. Translation of voice in the sentence.
4. Identification of the Proper Noun(s).
5. Use of Adjectives and Adverbs corresponding to the Nouns and Verbs.
6. Selection of proper words/synonyms (Lexical Choice).
7. Sequence of phrases and clauses in the translation.
8. Use of Punctuation Marks in the translation.
9. Fluency of translated text and translator's proficiency.
10. Maintaining the semantics of the source sentence in the translation.
11. Evaluating the translation of source sentence (With respect to syntax and intended meaning).

We have employed a five point scale to measure the quality of the translations. Table 11 provides the description of the scales.

| Score | Description |
|-------|-------------|
| 4 | Ideal |
| 3 | Perfect |
| 2 | Acceptable |
| 1 | Partially Acceptable |
| 0 | Not Acceptable |

Table 11: Interpretation of HEval on Scale 5

All the above features are scored using these scores. An average is computed on all the eleven scores which give us a single objective value. Figure 2 shows the working prototype that has been developed. Since the current human evaluation metrics were unable to provide the complete assessment of the dimensions of translation quality as they only focused on just one or two quality measures. Most of the human evaluation campaigns focused on limited features of evaluations. In these evaluations adequacy and fluency being the most important. This kind of evaluation is inappropriate as in practice a human translator does not translate a text just on superficial ratings. While translation, a translator has to look at many factors like gender being addressed in the text, the proper nouns being used, use of adjectives and adverbs etc. Then only a translator comes out with a good translation. Our metric intuitively captures this phenomenon and provides an objective score. Most of the MT systems might not be able to provide appropriate results for all the eleven features used in our metric, but they may provide good or acceptable results for some of the features. Judging an MT output on just one/two features/factors might deviate a human judge from making a sound judgment, as one judge, who has looked at all the features, may give a moderate rating to a translation whereas another judge, who might have overlooked some of the features, may give low rating to a translation. This causes an inter-annotator disagreement while evaluating MT Systems.

Moreover, if a human judge is asked to reevaluate a particular translation again, then he might not assign the same rating to the translation. This poses a serious problem while ascertaining quality of MT systems. HEval addresses these issues by clearly laying down parameters onto which an evaluation can be performed. This provides a crisp and repeatable assessment of MT outputs. Since, each feature clearing states what needs to be done and how to access the same; it reduces the possibility of assigning a low score to a good translation. Table 12 provides an example to justify this claim, where we show some of the good and bad translations and compare them with adequacy and fluency scores.

The scores of HEval Metric for these four translations are shown in table 13 along with adequacy and fluency scores. From the table it is evident that HEval metric can provide results in line with Human Adequacy and Fluency measures. Moreover, this metric can even justify its results as it provides more qualitative information about the evaluation. This metric also provide consistent scores when more than one judges evaluate a translation. All the judges came out with almost same scores for the translations produced by different engines. The good translations were consistently given a higher score whereas bad translations were given lower scores.

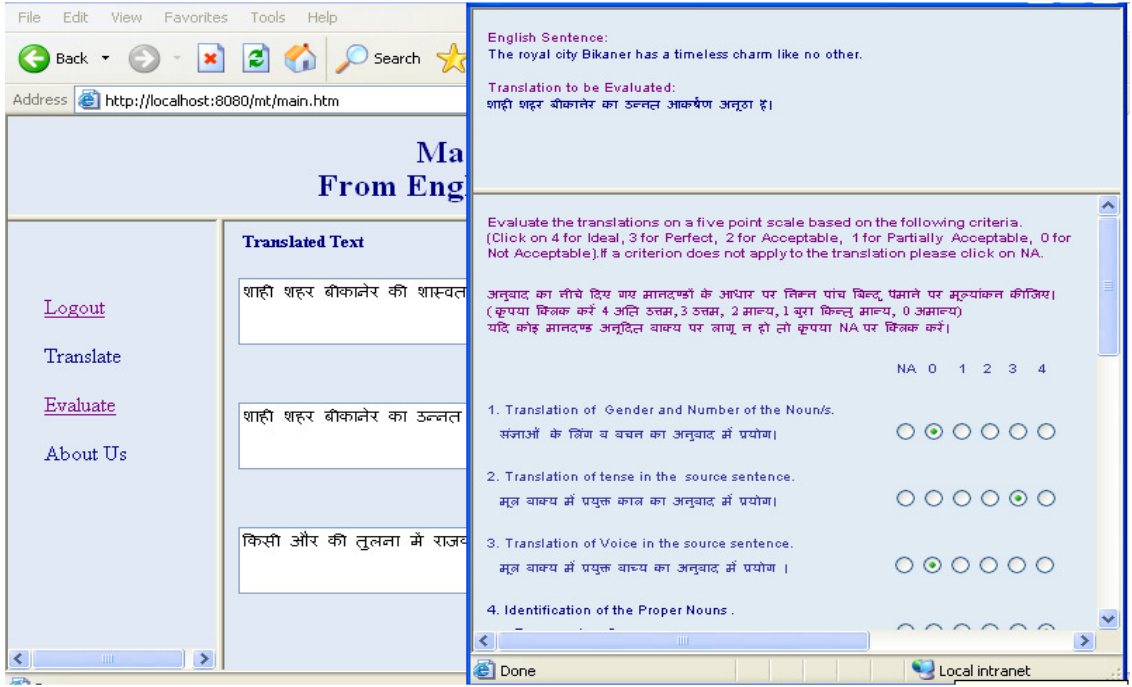

**Figure 2:** HEval: Subjective Human Evaluation Metric

| English | Most impressive are the black and white chessboard marble floor, the four tall minarets (40 m high) at the corners of the structure, and the majestic dome in the middle. |
|---|---|
| E1 | सबसे प्रभावशाली हैं काले और सफेद बिसात संगमरमर फर्श, चार लंबा मीनारों (40 मीटर ऊँची) संरचना है, और बीच में राजसी गुंबद के कोने में। |
| E2 | सबसे प्रभावशाली बिसात काले और सफेद संगमरमर का फर्श, संरचना के कोनों पर चार लंबा मीनारों (40 मीटर ऊंची), और बीच में राजसी गुंबद हैं. |
| E3 | ज्यादातर impressive हैं , काला और सफेद chessboard छतें मंजिल , के चार लंबे मीनारों (40 केंद्रीय high) पर के जनपथ के के structure , और के प्रकाशित गुम्बद के मध्य में है । |
| E4 | Most के के काला और सफेद संगमरमर chessboard floor , impressive हैं चार लंबे minarets के corners पर ( 40 मी . ऊँचा ) संरचना , मैजेस्टिक गुम्बद मध्य में और । |
| E5 | सबसे हैं के काले और सफेद संगमरमर मंजिल , चार ऊँची मीनारों ( मी . ऊँचे ) के कोने-कोने की संरचना , और के मैजेस्टिक गुम्बद के मध्य में 40 के पर है । |

Table 12: A Case of English-Hindi MT Evaluation (Sent # 150)

In order to understand how the metric came to a conclusion about a particular sentence, let us consider sentences of Engines E1 and E2. For feature 1 which checks for correct translation of gender, number and noun; the nouns available in the sentence are chessboard, marble, floor, minarets, m, high corners, dome and middle, gender is not an issue in this sentence as it is gender neutral, numbers available in this sentence are: four and 40. Among these, MT engine E1 could correctly translate only 8 out of 9 nouns while engine E2 could correctly translate all the nouns. Both the engines could very easily translate the number available in the text. As it is clearly seen that both the human judges gave four out of four to engine E2 and lesser score to engine E1.

| Engine | | 1 | 2 | 3 | 4 | 5 | 6 | 7 | 8 | 9 | 10 | 11 | Overall | Adq | Flu |
|--------|--------|---|---|---|----|---|---|---|---|---|----|----|---------|-----|-----|
| E1 | Human 1 | 3 | 4 | 4 | NA | 3 | 2 | 3 | 4 | 3 | 3 | 3 | 0.80 | 3 | 3 |
| | Human 2 | 2 | 4 | 4 | NA | 3 | 3 | 3 | 4 | 2 | 2 | 2 | 0.73 | | |
| E2 | Human 1 | 4 | 4 | 4 | NA | 4 | 3 | 4 | 4 | 3 | 3 | 3 | 0.93 | 4 | 4 |
| | Human 2 | 4 | 4 | 4 | NA | 4 | 4 | 4 | 4 | 4 | 3 | 3 | 0.95 | | |
| E3 | Human 1 | 2 | 2 | 2 | NA | 2 | 1 | 1 | 1 | 1 | 1 | 1 | 0.35 | 2 | 2 |
| | Human 2 | 2 | 2 | 2 | NA | 1 | 1 | 1 | 1 | 1 | 0 | 1 | 0.30 | | |
| E4 | Human 1 | 2 | 1 | 1 | NA | 1 | 0 | 1 | 3 | 1 | 1 | 1 | 0.30 | 2 | 2 |
| | Human 2 | 2 | 0 | 0 | NA | 1 | 0 | 1 | 3 | 1 | 0 | 1 | 0.23 | | |
| E5 | Human 1 | 2 | 2 | 2 | NA | 2 | 1 | 1 | 2 | 1 | 1 | 2 | 0.4 | 2 | 3 |
| | Human 2 | 2 | 2 | 2 | NA | 2 | 1 | 1 | 2 | 1 | 1 | 2 | 0.4 | | |

Table 13: Results of Human Evaluation for Case of Table 10

For feature 2 which checks for the translation of the tense in the source sentences. Both the engines could easily retain the tense in the translated text, thus they both have maximum score for this feature. For feature 3 which translates the voice; both the engines have retained the voice. For feature 4 where translation of proper noun is considered; since there was no proper noun in the text this feature was not judged and a 'NA' was given to the score. For feature 5 which checks for the translation of adjectives and adverbs with respect to nouns and verbs respectively; we have a pair of adverb and adjective (most impressive) with the verb "are", adjectives "black", "white" with noun "chessboard marble floor" and adjective "majestic" with noun "dome". Engine E2 could translate these very easily but engine E1 had trouble while managing the sequence for "black and while chessboard marble floor". Since it could not correctly translate this part, it was adjudged to be somewhat correct and was give a score of 3 out of 4. For feature 6, engine E2 was able to translate most of the words correctly and thus was give a score of 4 and 3 by judge1 and judge2 accordingly. For engine E1, some of the words were not properly translated thus a low score was given to it. For feature 7, Engine E2 preserved most of the sequences of the nouns, verbs and helping verbs in the translation thus was given 4 out of 4 by both the judges. Engine E1 had trouble doing this, thus was given a low score again. For feature 8, both the engines preserved the punctuations in the sentence, thus both got the same score. For feature 9, engine E2 could very well preserve the significant part of the translation of the source sentence, thus got 4 out of 4. Since this was not the case with engine E1, it scored 3 and 2 by judge1 and judge2 respectively. For feature 10, Engine E2 was able to preserve most of the semantics of the source sentence while Engine E1 was not able to capture some of the semantics, thus they were adjudged accordingly. For feature 11, Engine E2 was able to translate most of the sentence correctly thus was given a moderate rating while engine E1 was able to capture most of syntax but not complete semantics, thus was given low rating as compared to engine E2.

For the final calculation of the results of the MT engines, each individual score was added, leaving aside the ones which were not applicable. In our case, feature 4 was not applicable, thus it was left and rest of the scores were computed. The final score was calculated using equation 1.

$$Final\ Score\ Engine_i = \frac{\sum_{i=1}^{11} feature\ score_i}{4 \times (\#applicable\ features)} \qquad (1)$$

Here we have only considered the features that are applicable onto a sentence and have not considered the ones which are not applicable as they do not contribute to the final score computation and while considering it in division would be highly inappropriate.

## 5. EVALUATION AND RESULTS

We evaluated 1300 sentences onto the MT engines which were divided into 13 documents of 100 sentences each. Table 14 shows the average scores of 100 sentences for each document. For human judge1, out of the 13 documents engine E1 could score the highest in five documents and for the rest of the eight documents, engine E2 scored the highest. For human judge2, engine E1 scored highest score in four documents and engine E2 score the highest in the remaining documents. While considering only engines E3, E4 and E5; E5 scored the highest is almost all the documents for both the judges. This trend was repeatable when we computed the scores for entire systems. For this we computed the averages of all 1300 sentence scores. Table 15 shows the system level average scores of the MT engines. Here too, engine E2 scored the highest average score. While considering our limited corpus systems, E5 again scored highest among all the three engines.

|  |  | E1 | E2 | E3 | E4 | E5 |
|---|---|---|---|---|---|---|
| Doc1 | Human1 | **0.5408** | 0.535 | 0.4338 | 0.4455 | 0.428 |
|  | Human2 | **0.5554** | 0.543 | 0.4085 | 0.4322 | 0.4275 |
| Doc2 | Human1 | 0.6311 | **0.671** | 0.2733 | 0.2788 | 0.4918 |
|  | Human2 | 0.5741 | **0.6214** | 0.2366 | 0.2354 | 0.3936 |
| Doc3 | Human1 | 0.624 | **0.6335** | 0.4993 | 0.3455 | 0.5008 |
|  | Human2 | **0.5748** | 0.5704 | 0.4727 | 0.3641 | 0.468 |
| Doc4 | Human1 | 0.647 | **0.6913** | 0.2858 | 0.2865 | 0.4373 |
|  | Human2 | 0.6366 | **0.6445** | 0.3184 | 0.3162 | 0.4515 |
| Doc5 | Human1 | 0.5413 | **0.5933** | 0.4285 | 0.3688 | 0.4865 |
|  | Human2 | 0.6469 | **0.6602** | 0.495 | 0.4546 | 0.5392 |
| Doc6 | Human1 | 0.6743 | **0.7013** | 0.454 | 0.4458 | 0.4743 |
|  | Human2 | 0.6069 | **0.6245** | 0.3822 | 0.3639 | 0.4337 |
| Doc7 | Human1 | **0.6938** | 0.6028 | 0.353 | 0.292 | 0.4333 |
|  | Human2 | 0.6265 | **0.6468** | 0.4793 | 0.4035 | 0.489 |
| Doc8 | Human1 | 0.7223 | **0.7363** | 0.495 | 0.4425 | 0.554 |
|  | Human2 | 0.5325 | **0.539** | 0.3693 | 0.2605 | 0.4308 |
| Doc9 | Human1 | **0.6793** | 0.6428 | 0.3403 | 0.3995 | 0.4633 |
|  | Human2 | **0.6889** | 0.6356 | 0.4257 | 0.4015 | 0.4655 |
| Doc10 | Human1 | **0.5388** | 0.5298 | 0.4425 | 0.3705 | 0.4723 |
|  | Human2 | 0.5644 | **0.584** | 0.4013 | 0.3342 | 0.4544 |
| Doc11 | Human1 | 0.596 | **0.6753** | 0.3388 | 0.2455 | 0.503 |
|  | Human2 | 0.6006 | **0.6599** | 0.3464 | 0.2644 | 0.4609 |
| Doc12 | Human1 | **0.6443** | 0.6338 | 0.321 | 0.2718 | 0.4068 |
|  | Human2 | **0.618** | 0.6091 | 0.3411 | 0.3234 | 0.4016 |
| Doc13 | Human1 | 0.636 | **0.6478** | 0.4025 | 0.384 | 0.5133 |
|  | Human2 | 0.6581 | **0.6612** | 0.3903 | 0.3748 | 0.5226 |

Table 14: Document Level Scores for MT Engines

|          | E1     | E2     | E3     | E4    | E5     |
|----------|--------|--------|--------|-------|--------|
| **Human1** | 0.6284 | **0.638** | 0.3898 | 0.352 | 0.4742 |
| **Human2** | 0.6201 | **0.6247** | 0.3933 | 0.356 | 0.464 |

Table 15: System Level Scores for MT Engines

To check the results of the metric at sentence level, we computed the ranks for all the sentences. Table 16 shows the highest ranks scored by each engine. Here too, engine E2 scored the highest scores for most of the sentences. For limited corpus systems, E5 scored the highest rank for most of the sentence level evaluations. Table 17 shows the results of the same. Figure 3 and 4 show these trends.

| Engine | Human1 | Human2 |
|--------|--------|--------|
| E1 | 454 | 388 |
| E2 | 672 | 529 |
| E3 | 45 | 73 |
| E4 | 33 | 55 |
| E5 | 96 | 255 |

Table 16: Sentence Level Ranking for MT Engines

| Engine | Human1 | Human2 |
|--------|--------|--------|
| E3 | 294 | 341 |
| E4 | 266 | 324 |
| E5 | 740 | 635 |

Table 17: Sentence Level Ranking for Limited Corpus MT Engines

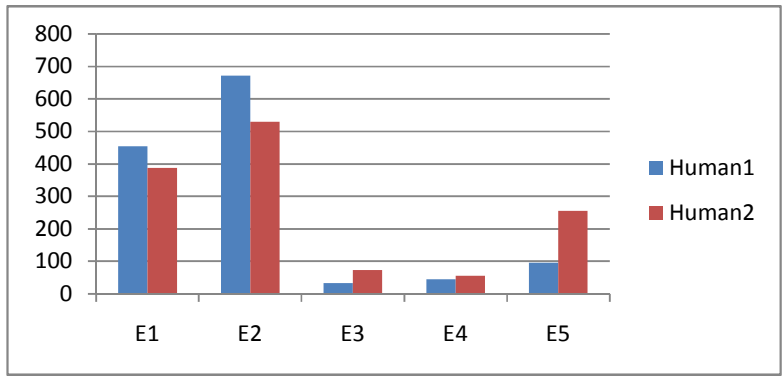

Figure 3: Sentence Level Ranking for MT Engines

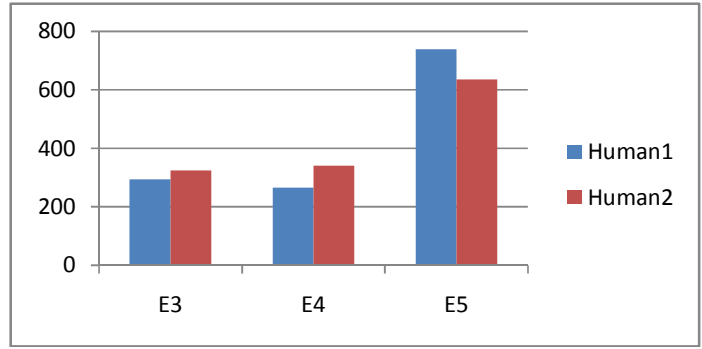

Figure 4: Sentence Level Ranking for Limited Corpus MT Engines

While considering inter-annotator agreement, we considered the highest rank given to an MT output by both the judges. Table 18 shows the results of this study. Out of the 1300 sentences, 842 times, both the judges gave same high rank to the same MT Engine's output. This scored around 67% agreement percentage which according to us is a good score. While considering MT engines with limited corpus, we found that 804 times the highest rank was given to the same engine by both the human judges. This attributed to around 61% agreement percentage which is again a very good agreement percentage. While considering engine wise inter-annotator agreement. We counted the no. of times the same rank was given to a sentence of an engine by both the judges. Table 19 shows the results of this study. In all the cases it was found that the trend is repeatable.

| | Combined | | Limited Corpus | |
|---|---|---|---|---|
| | Agreement | Percentage | Agreement | Percentage |
| Inter-Annotator Agreement | 842 | 64.77% | 804 | 61.85% |

Table 18: Inter-Annotator Agreement for Highest Rank

| Engine | Agreement | Percentage |
|---|---|---|
| E1 | 849 | 65.30% |
| E2 | 885 | 68.07% |
| E3 | 959 | 73.77% |
| E4 | 967 | 74.38% |
| E5 | 1067 | 82.07% |

Table 19: Engine wise Inter-Annotator Agreement

We have also compared the results produced by this metric with fluency and adequacy measures which are considered to be popular human evaluation measures. We used only one judge to evaluate the MT engines with fluency and adequacy measures. The results of this evaluation was compared with the ones produced by both judges who evaluated the engines using HEval metric. Table 20 shows the results of correlation between human adequacy measure for scale 5 and HEval metric. At 95% confidence interval, both the judges' results showed positive trend for all the MT engines, thus strengthening our claim that this metric can be a substitute for the Human adequacy measure.

| Engine | Human1 | Human2 |
|--------|--------|--------|
| E1 | 0.0705 | 0.0681 |
| E2 | 0.0851 | 0.1492 |
| E3 | 0.079 | 0.048 |
| E4 | 0.027 | 0.0104 |
| E5 | 0.1294 | 0.0493 |

Table 20: Correlation between Human Adequacy Measure Scale 5 and HEval Metric at 95% Confidence Level

| Engine | Human1 | Human2 |
|--------|--------|--------|
| E1 | 0.0685 | 0.0484 |
| E2 | 0.0686 | 0.1055 |
| E3 | 0.0549 | 0.0594 |
| E4 | 0.0518 | 0.0667 |
| E5 | 0.0353 | 0.0165 |

Table 21: Correlation between Human Fluency Measure Scale 5 and HEval Metric at 95% Confidence Level

Table 21 shows the results of correlation between human fluency measure for scale 5 and HEval metric. At 95% confidence interval, both the judges' results showed positive trend as well. This further justifies our claim that this metric can act as a substitute for human adequacy and fluency measure.

# 6. CONCLUSION

In this paper, we showed the working of a new human evaluation metric. We showed the design of the metric along with the scoring criteria and the mechanism for scoring the translations using the criteria with an example. We have also compared the results of this metric with the popular human evaluation measures (Adequacy and Fluency) which have been implemented in almost all the evaluation campaigns. We also correlated the results of our metric with these two measures. The metric showed significant correlation with both type of MT Engines.

We also analyzed the inter-annotator agreement and found that this metric can produce better agreements between the results of the multiple judges providing judgements for the same set of sentences. This can be considered as an advantage of this metric. As human evaluation is very subjective, the results produced by multiple judges often do not have an agreement. We also analyzed the evaluation results produced at different levels and found that at all three levels (system, document and sentence level) the results were consistent i.e. engine E2 was adjudged as the best MT engine throughout, at the same time, engine E4 was adjudged as the engine which could not do so well. Moreover, although the evaluation process was subjective in nature, the metric can very easily provide answers as to why a particular translation was termed as good or bad. Moreover it can also provide micro level details about the evaluation process for the translation like if the translation of nouns was proper or not, translation of adverbs and adjectives with respect to verbs and nouns was proper or not etc. This information can help MT development team understand the short comings of their MT engine, so that they may focus on the enhancement of that particular area.

## AUTHORS

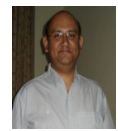

Dr. Nisheeth Joshi is an Assistant Professor at Banasthali University. His primary area of research is Machine Translation Evaluation. He has designed and developed some evaluation metrics in Indian Languages. Besides this he is also very actively involved in the development of MT engines for English to Indian Languages. He has several publications in various journals and conferences and also serves on the Programme Committees and Editorial Boards of several conferences and journals.

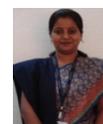

Mrs. Iti Mathur is an Assistant Professor at Banasthali University. Her primary area of research is Computational Semantics and Ontological Engineering. Besides this she is also involved in the development of MT engines for English to Indian Languages. She is one of the experts empanelled with TDIL Programme, Department of Electronics and Information Technology (DeitY), Govt. of India, a premier organization which foresees Language Technology Funding and Research in India. She has several publications in various journals and conferences and also serves on the Programme Committees and Editorial Boards of several conferences and journals.

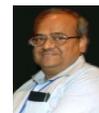

Dr. Hemant Darbari, Executive Director of the Centre for Development of Advanced Computing (CDAC) specialises in Artificial Intelligence System. He has opened new avenues through his extensive research on major R&D projects in Natural Language Processing (NLP), Machine assisted Translation (MT), Information Extraction and Information Retrieval (IE/IR), Speech Technology, Mobile computing, Decision Support System and Simulations. He is a recipient of the prestigious "Computerworld Smithsonian Award Medal" from the Smithsonian Institution, USA for his outstanding work on MANTRA-Machine Assisted Translation Tool which is also a part of "The 1999 Innovation Collection" at National Museum of American History, Washington DC, USA.

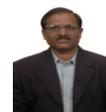

Mr. Ajai Kumar, Additional Director of Applied AI Group, Centre for Development of Advanced Computing, Pune has extensive experience in development and deployment of AI systems. He has been instrumental in completing many AI projects that are deployed in various state and central government departments and various defence establishments. He has extensive experience in R&D projects in Natural Language Processing (NLP), Machine assisted Translation (MT), Information Extraction and Information Retrieval (IE/IR), Speech Technology.